\documentclass[letterpaper, 10 pt, conference]{ieeeconf}  

\IEEEoverridecommandlockouts                              

\usepackage{times}
\usepackage{graphicx,stfloats}
\DeclareGraphicsExtensions{.pdf,.png,.jpg}
\usepackage{amsmath}
\usepackage{amssymb}  
\usepackage{makecell}
\usepackage{booktabs}
\usepackage{multirow}
\usepackage{multicol}
\usepackage{here}
\usepackage{kotex}
\usepackage{soul, color}
\usepackage{subcaption} 
\usepackage{anyfontsize} 
\usepackage{hyperref}
\usepackage{booktabs}  
\usepackage{cuted}
\usepackage[font=small,labelfont=bf,tableposition=top]{caption}
\usepackage{etoolbox}
\usepackage{cleveref}
\usepackage{comment}
\usepackage{xcolor}
\title{\LARGE \bf
Bayesian NeRF: Quantifying Uncertainty with Volume Density for Neural Implicit Fields}

\author{Sibaek Lee, Kyeongsu Kang, Seongbo Ha and Hyeonwoo Yu
 \thanks{Sibaek Lee, Kyeongsu Kang, Seongbo Ha and Hyeonwoo Yu are with the Department of Intelligent Robotics, Sungkyunkwan University, Suwon, South Korea. {\tt\small \{lmjlss, thithin0821, sobo3607, hwyu\}@skku.edu}}
 \thanks{
 The code is available at: \href{https://github.com/Lab-of-AI-and-Robotics/Bayesian_NeRF.git}{https://github.com/Lab-of-AI-and-Robotics/Bayesian\_NeRF}}
}

\begin{document}

\maketitle
\thispagestyle{empty}
\pagestyle{empty}

\begin{abstract}


We present a Bayesian Neural Radiance Field (NeRF), which explicitly quantifies uncertainty in the volume density by modeling uncertainty in the occupancy, without the need for additional networks, making it particularly suited for challenging observations and uncontrolled image environments. NeRF diverges from traditional geometric methods by providing an enriched scene representation, rendering color and density in 3D space from various viewpoints. However, NeRF encounters limitations in addressing uncertainties solely through geometric structure information, leading to inaccuracies when interpreting scenes with insufficient real-world observations. While previous efforts have relied on auxiliary networks, we propose a series of formulation extensions to NeRF that manage uncertainties in density, both color and density, and occupancy, all without the need for additional networks. In experiments, we show that our method significantly enhances performance on RGB and depth images in the comprehensive dataset. 
Given that uncertainty modeling aligns well with the inherently uncertain environments of Simultaneous Localization and Mapping (SLAM), we applied our approach to SLAM systems and observed notable improvements in mapping and tracking performance. These results confirm the effectiveness of our Bayesian NeRF approach in quantifying uncertainty based on geometric structure, making it a robust solution for challenging real-world scenarios.
\end{abstract}

\section{\textbf{INTRODUCTION}}
The advent of Neural Radiance Fields (NeRF) \cite{mildenhall2021nerf} has significantly advanced the field of novel view synthesis, allowing for the continuous synthesis of views from given images at unseen positions and orientations. As opposed to traditional geometric approaches \cite{schonberger2016structure, seitz2006comparison}, NeRF leverages a learned neural network model to predict color and density in 3D space from coordinates and view directions, thus better handling complex scenes, intricate textures, and lighting variations. Subsequent research has propelled its real-world application in robotics \cite{rosinol2023nerf, wang2023co}, virtual reality \cite{deng2022fov}, digital twins \cite{liu2023visualization}, and autonomous driving \cite{wu2023mars, hu2023pc}, underscoring the importance of real-time performance and efficiency when data is limited. 
However, advancements are hindered by challenges like accurately predicting scenes in unobserved views with limited data availability \cite{xiang2015data}. This is inherently challenged by the limitations of the sensor's field of view (FoV) and physical occlusions, such as buildings or other vehicles, frequently result in gaps in observable data. Moreover, the intrinsic nature of the real-world ensures that sensor data is invariably accompanied by errors \cite{thrun2002probabilistic}. To overcome these challenges, it is necessary to incorporate uncertainty consideration and this approach is crucial for precise interpretation of and adaptation to the variabilities encountered in real-world environments \cite{blundell2015weight, kendall2017uncertainties, abdar2021review}.

Given the importance of considering uncertainty in neural radiance fields, various studies have focused on addressing the variability of color in space. \cite{martin2021nerf} tackled photometric variations to deal with images captured in various environments, while \cite{pan2022activenerf} used data selection strategies for enhanced learning. They incorporate an L1 regularization term to mitigate transient density issues. 
However, considering uncertainty in color presents several critical drawbacks. Firstly, it fails to address uncertainties in other sensor data such as density, which are essential for a comprehensive understanding of 3D spaces. Moreover, predicting pixel variance does not fully capture the uncertainty in the scene's radiance field and can lead to suboptimal results. 
In response to these challenges, \cite{shen2021stochastic, shen2022conditional} suggested employing an additional network to consider the density uncertainty. However, the use of an extra network is inefficient for real-time applications and does not guarantee accurate estimations due to the approximation of the probability distribution.

\begin{figure}[tb]
  \centering
  \includegraphics[width=0.46\textwidth]{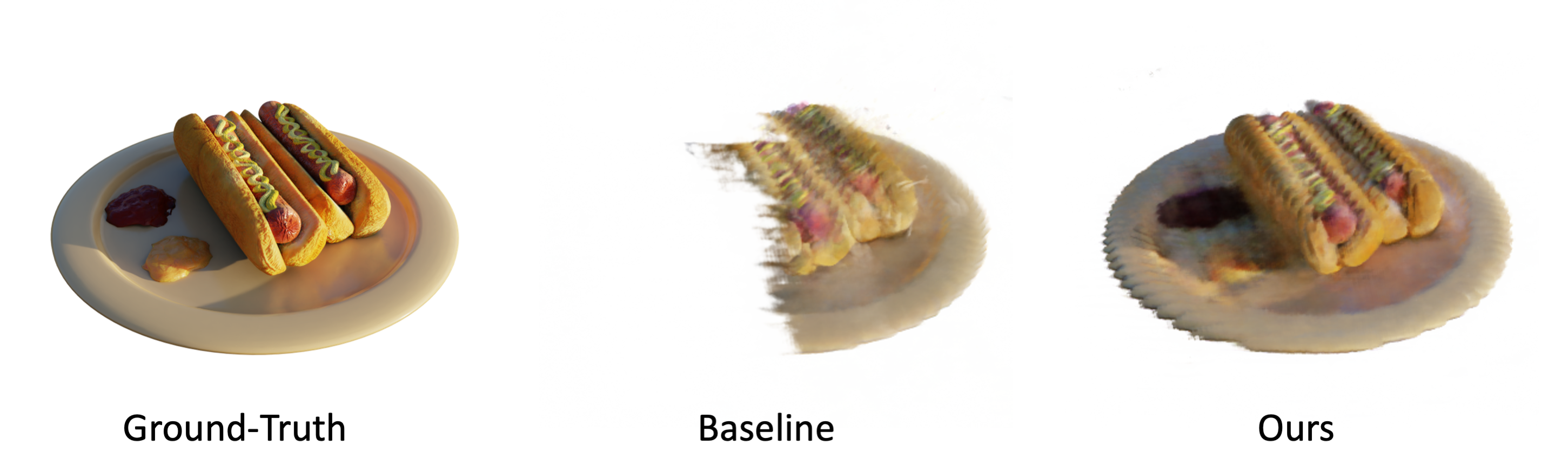}
  \caption{\textbf{Quantitative results.} Ground Truth (Left), baseline model prediction (Middle), our method's prediction (Right). All models were trained on 4 images from the NeRF dataset and predict unobserved viewpoints. } 
  \label{fig:title_figure}
  \vspace{-15pt}
\end{figure}

We introduce a series of methodical approaches that allow us to incorporate density-related uncertainty into the NeRF model without altering its network structure. This methodology not only explicitly addresses density uncertainty but also renders the model suitable for real-time applications by leveraging Bayesian techniques to manage and interpret the uncertainties inherent in the data. 
Our approaches improves rendering in unobserved views, as shown in \cref{fig:title_figure}, where training data from limited viewpoints lead to suboptimal performance. Incorporating a Bayesian framework enhances the model's capacity to handle unseen data uncertainties, boosting rendering quality from various angles.
Furthermore, we address the limitations of sensors that do not capture color data, such as Lidar and Thermal Imaging Cameras, which make traditional uncertainty methods ineffective.
By focusing on density as a universal attribute detectable across different sensor types, our method extends the applicability of NeRF models beyond the realm of RGB cameras. Integrating uncertainty across different sensors broadens NeRF's flexibility and significantly improves performance in real-world scenarios, demonstrating the robustness and adaptability of our enhanced framework.

\section{\textbf{Related work}}
\label{sec:related}

\subsection{\textbf{Novel-view synthesis and NeRF}}
The field of novel-view synthesis has seen remarkable advancement as computer vision technologies have evolved. Initially, traditional methods were concentrated on understanding the structure of 3D scenes through camera poses, utilizing techniques such as structure-from-motion (SFM) \cite{schonberger2016structure} and Multi-View Stereo (MVS) \cite{seitz2006comparison}. These methods, based on geometric approaches, aimed to synthesize images from novel viewpoints by accurately modeling 3D structures. The advent of NeRF marked a significant leap forward, enabling the creation of realistic renderings from a smaller set of images through 3D representations learned by neural networks. NeRF estimates volume density and color information based on 3D coordinates and viewing directions, with its network output facilitating image generation via a rendering function. These images reflect observations within the modeled space, informed by the direction of the camera view \cite{martin2021nerf, pan2022activenerf, shen2022conditional}.
Efforts have since been directed at addressing the original NeRF's limitations, spurring a variety of enhancements. For instance, changes to encoding methods have sped up the learning process \cite{muller2022instant, barron2023zip, hu2023tri}, while large-scale scenes are now depicted using multiple implicit neural networks \cite{yu2021pixelnerf, yang2023freenerf}. Additionally, the introduction of new priors has facilitated learning from sparse data \cite{tancik2022block, Turki_2022_CVPR}. NeRF's application scope has also expanded, encompassing areas such as scene editing \cite{wang2022clip, yuan2022nerf}, converting text to 3D models \cite{poole2022dreamfusion, liu2023zero}, and enhancing visual scene-based SLAM technologies \cite{sucar2021imap, zhu2022nice, kong2023vmap, rosinol2023nerf, yang2022vox}, demonstrating its versatility and potential in various domains.


\subsection{\textbf{Uncertainty estimation in NeRF}}
Estimating uncertainty is a well-known and important issue in deep learning. Similarly, in NeRF, considering uncertainty is crucial for reducing errors stemming from sensor noise, sensor field of view (FOV), occlusions, and other factors, enabling the representation of space with reliability and robustness. Due to this importance, research efforts are underway to integrate uncertainty estimation into existing NeRF-based techniques. For instance, \cite{martin2021nerf} accurately render scenes from images taken in various environments, \cite{pan2022activenerf} select data for subsequent learning, and \cite{lee2023just} consider uncertainty in color to measure the reliability of generated images. While these approaches explicitly estimate uncertainty and do not require more extensive training than baseline models, they have the drawback of not considering uncertainty for volume density. Other recent works \cite{goli2023bayes} represent rendering cleanup by measuring uncertainty in trained models. \cite{sunderhauf2023density} proposes a method that utilizes an ensemble of multiple NeRF models to quantify uncertainty, which can be applied to next view selection. However, these approaches can only capture epistemic uncertainty and cannot capture inherent aleatoric uncertainty in the data.
Addressing volume density uncertainty, \cite{shen2021stochastic, shen2022conditional} presents a method that employs an additional network to measure depth uncertainty.
This method is difficult to apply in real-time situations due to additional network usage. To address this challenge, we introduce a series of methods based on Bayesian approaches that explicitly estimate volume density uncertainty, tackling the complex issue of depth uncertainty and effectively predicting unobserved views. Our methodologies formulate uncertainty for volume density and offer ways to quantify uncertainty without the need for additional estimation networks.

\section{\textbf{Methodology}}
\label{sec:methodology}

\subsection{\textbf{Background}}
NeRF is a coordinate-based neural representation of a 3D volumetric scene \cite{mildenhall2021nerf}. The goal is to find a continuous non-linear regression $F_\Theta : (\mathbf{x},\mathbf{d}) \rightarrow (\mathbf{c}, \rho)$, where $\Theta$ represents the network parameters, $\mathbf{x}= (x,y,z)$ are the spatial 3D coordinates, and $\mathbf{d} = (\theta, \phi)$ are the viewing directions of the camera. The output of $F_\Theta$ estimates the color $\mathbf{c}$ and density $\rho$ at a given point in the neural field.
To render an image, the color of each pixel is determined by integrating colors along a ray in the neural field. The expected image color $C(\mathbf{r})$ for a camera ray $\mathbf{r}(t) = \mathbf{o} + t\mathbf{d}$ starting from the camera origin $\mathbf{o}$ with near and far bounds $t_n$ and $t_f$ is given by:
\begin{align}
    C(r) &= \int_{t_n}^{t_f} T(t)\rho(\mathbf{r}(t))
    \mathbf{c}(\mathbf{r}(t),\mathbf{d})dt\text{,\;}
    \label{eq:rendering function}
\end{align}
where $T(t)$ is the accumulated transmittance along the ray. For practical implementation, the integral is approximated by sampling $N$ discrete points along the ray at positions $t_i$, spaced by $\delta_i = t_{i+1} - t_{i}$. This discretization maintains a continuous scene representation, and the rendering equation becomes:
\begin{align}
    \nonumber
    \hat{C}(r) 
    & = \sum_i^N 
    \underbrace{T_i (1-\exp(-\delta_i \rho_i))}_{\alpha_i}
    c_i
    \\
    & = \sum_i^N 
    \alpha_i
    c_i
    \text{,\; where } 
    T_i = \exp\left(-\sum^{i-1}_{j=1} \delta_j \rho_j \right) \text{.}
\label{eq:discrete rendering function}
\end{align}

\subsection{\textbf{Uncertainty Estimation}}
Training with limited observations presents a challenge in accurately predicting unobserved parts of the 3D scene. To effectively tackle the problem, it is essential to adopt the Bayesian learning \cite{kendall2017uncertainties} for considering the observation and estimation uncertainty.
For example, \cite{martin2021nerf, pan2022activenerf} simply assume that only the color $\mathbf{c}$ follows a Gaussian distribution $\mathbf{c}\sim\mathcal{N}\left(\mu_{\mathbf{c}}, \sigma_{\mathbf{c}}^2\right)$, while treating density $\rho$ and transmittance $T$ as constants.
As in ~\cref{eq:discrete rendering function}, since the cumulated color $C(\mathbf{r})$ along the ray $\mathbf{r}$ is a linear combination of Gaussian random variable $\mathbf{c}$, it also follows the Gaussian distribution
\begin{align}
        C\left(\mathbf{r}\right)
    \sim
    \mathcal{N}\left(
        \sum^{N_s}_{i=1}\mu_i \alpha_i,
        \sum^{N_s}_{i=1}\sigma^2_i \alpha_i^2
    \right)
    .    
\end{align}
This assumption makes it easy to model the variability in color predictions, facilitating the estimation of color uncertainty and its impact on the rendered scene.
However, this color-only approach is limited by considering only RGB images, making it impossible to extend to other sensor values such as depth or intensity. To address the general uncertainty in estimating 3D space from sensors, one can consider incorporating a density field in neural representation.
Unfortunately, unlike color-only case, when we assume that density $\rho$ follows a Gaussian distribution $\rho\sim\mathcal{N}\left(\mu_d, \sigma_d^2\right)$ the problem becomes highly complicated.
Since 
$T_i=\exp
\left(
    -\sum^{i-1}_{j=1} 
        \delta_j \rho_j
\right)$ as in ~\cref{eq:discrete rendering function}, we have $T_i \sim Lognormal\left(\mu_{T_i}, \sigma_{T_i}^2\right)$ where $\mu_{T_i} = -\sum^{i-1}_{j=1}\delta_j\mu_j$ and $\sigma_{T_i}^2 = \sum^{i-1}_{j=1}\delta^2_j\sigma^2_j$.
Meanwhile, $\alpha$ in ~\cref{eq:discrete rendering function} can be rewritten as:
\begin{align}
    \nonumber
    \alpha_i &= \exp
    \left( -\sum^{i-1}_{j=1} \delta_j \rho_j \right)
    \left( 1 - \exp \left( -\delta_i \rho_i \right) \right)
    \\
    \nonumber
    &= \exp \left( -\sum^{i-1}_{j=1} \delta_j \rho_j \right)
    -
    \exp \left( -\sum^{i}_{j=1} \delta_j \rho_j \right)
    \\
    &= T_i - T_{i+1}
    \text{,}
    \label{eq:alpha}
\end{align}
which is the difference of two lognormal random variables $T_i$ and $T_{i+1}$.
It is known to be challenging to have the closed-form distribution of the difference of two lognormal random variables, even the two random variables are independent and thus uncorrelated \cite{lognorm,lo2012sum}.
Therefore, even if we assume that the color $\mathbf{c}$ in ~\cref{eq:discrete rendering function} is a constant, $C(\mathbf{r})$ still follows an intractable distribution since $C(\mathbf{r})$ is a linear combination of $\alpha$.
To model this complex distribution, recent works such as \cite{shen2021stochastic, shen2022conditional} utilize generative models and additional network structures, resulting in complex frameworks and challenging learning tasks.
In the following, we introduce a series of our analytic but simple and effective approaches with consideration of density uncertainty without the need of additional network structures or trainings.

\subsubsection{Gaussian Approximation of Density Uncertainty}
We propose a method to optimize the NeRF model by considering density uncertainty through a series of assumptions.
Using $\alpha_i$ represented in ~\cref{eq:alpha}, expected color $C\left(\mathbf{r}\right)$ and expected depth $D\left(\mathbf{r}\right)$ of the ray $\mathbf{r}$ can be derived as \cite{tao2023lidar, zhang2023nerf, huang2023neural, malik2023transient}:
\begin{align}
    C\left(\mathbf{r}\right)
    & =
    \sum^{N_s}_{i=1} \mathbf{c}_i \alpha_i  
    \text{,} \label{eq:color} \\
    D\left(\mathbf{r}\right)
    &= \sum^{N_s}_{i=1} d_i \alpha_i = 
    \sum^{N_s}_{i=1} \frac{t_{i+1} + t_i}{2} \alpha_i
    \label{eq:depth}
    \text{.}
\end{align}
Here, let us make a slightly strong assumption regarding $\alpha$, such that $\sum\delta_j \rho_j \ll 1$.
In other words, we assume that the intervals between samples are very narrow.
Furthermore, neural fields are generally sparse, and the volume density is forced to take values less than $1$ using activation functions such as sigmoid.
Then we can approximately represent \cref{eq:alpha} as the following:
\begin{align}
    \alpha_i
    \approx
    \left( 1-\sum^{i-1}_{j=1} \delta_j \rho_j \right)
    -
    \left( 1-\sum^{i}_{j=1}  \delta_j \rho_j \right)
    = \delta_i \rho_i
    \text{.}
    \label{eq:approximation_alpha}
\end{align}
Now let $k$'th volume density $\rho_k$ follows the Gaussian distribution $\rho_k\sim\mathcal{N}\left(\mu_k, \sigma_k^2\right)$ and the densities are i.i.d.. Assume that the color $\mathbf{c}$ is a constant.
Substituting \cref{eq:approximation_alpha} into \cref{eq:color}, we approximately have the rendered color as $C\left(\mathbf{r}\right)\approx\sum^{N_s}_{i=1} \mathbf{c}_i \delta_i \rho_i$.
As the rendered color $C\left(\mathbf{r}\right)$ is now a linear combination of Gaussian random variables, its distribution can be written as:
\begin{align}
C\left(\mathbf{r}\right)
\sim
\mathcal{N}\left(
    \sum^{N_s}_{i=1}\mathbf{c}_i\delta_i\mu_i,
    \sum^{N_s}_{i=1}\mathbf{c}^2_i\delta^2_i\sigma^2_i
\right) \text{,}
\end{align}
and we can obtain the approximated solution $\{\left( \hat{\mu}_i, \hat{\sigma}_i\right) \}$ of the volume density by minimizing the following negative log-likelihood:
\begin{align}
    \ln \sum^{N_s}_{i=1}\mathbf{c}^2_i\delta^2_i\sigma^2_i
    +
    \frac
    {\left( C\left(\mathbf{r}\right) - \sum^{N_s}_{i=1}\mathbf{c}_i\delta_i\mu_i \right)^2}
    {\sum^{N_s}_{i=1}\mathbf{c}^2_i\delta^2_i\sigma^2_i}
    \text{.}
    \label{eq:gaussian_approximaged_MLE_color}
\end{align}
Similarly, we can obtain the approximated solution $\{\left( \hat{\mu}_i, \hat{\sigma}_i\right) \}$ of $D\left(\mathbf{r}\right)$ by minimizing the following negative log-likelihood:
\begin{align}
    \ln \sum^{N_s}_{i=1}d^2_i\delta^2_i\sigma^2_i
    +
    \frac{
    \left(
        D\left(\mathbf{r}\right) - \sum^{N_s}_{i=1}d_i\delta_i\mu_i
    \right)^2
    }{
    \sum^{N_s}_{i=1}d^2_i\delta^2_i\sigma^2_i   
    }
    \text{.}
    \label{eq:gaussian_approximaged_MLE_depth}
\end{align}
By using \cref{eq:gaussian_approximaged_MLE_color} or \cref{eq:gaussian_approximaged_MLE_depth}, now we can consider the uncertainty of both color or depth, or other sensor information, by utilizing the density uncertainty.

\subsubsection{Gaussian Approximation of Density and Color Uncertainty}
In addition to the above method, we also propose a method that considers both color and density by assuming that color follows a Gaussian distribution, rather than being constant.
Specifically, we assume that $\mathbf{c}_k\sim\mathcal{N}\left(\mu_{\mathbf{c}_k}, \sigma_{\mathbf{c}_k}^2\right)$, then \cref{eq:color} becomes the summation of the product of two Gaussian random variables, $\rho$ and $\mathbf{c}$. 
While the distribution of this product is infeasible to compute explicitly, under certain conditions, specifically when the means of the Gaussian distributions are significantly larger than their standard deviations (i.e., $\mu_k\gg\sigma_k$ and $\mu_{\mathbf{c}_k}\gg\sigma_{\mathbf{c}_k}$), the product can be approximated by a Gaussian distribution \cite{seijas2012approach}.
Practically, this means that the relative variability of the random variables is low, and their distributions are sharply peaked around the mean. This allows us to approximate the product $\rho_k \mathbf{c}_k$ as another Gaussian distribution with the following mean and variance:
\begin{align}
\nonumber
    \sigma_k \mathbf{c}_k
    \sim
    \mathcal{N}
    \left(
        \mu_k \mu_{\mathbf{c}_k},
        \sigma^2_k\mu^2_{\mathbf{c}_k} + \sigma^2_{\mathbf{c}_k}\mu^2_k + \sigma^2_k\sigma^2_{\mathbf{c}_k}
    \right)
    \text{.}
\end{align}
Consequently, the distribution of rendered color $C\left(\mathbf{r}\right)$ can be approximated as:
\begin{align}
C\left(\mathbf{r}\right)
\sim
\mathcal{N}\left(
    \sum^{N_s}_{i=1}\delta_i\mu_i\mu_{\mathbf{c}_i},
    \sum^{N_s}_{i=1}\delta^2_i
    \left(
        \sigma^2_i\mu^2_{\mathbf{c}_i} + \sigma^2_{\mathbf{c}_i}\mu^2_i + \sigma^2_i\sigma^2_{\mathbf{c}_i}
    \right)
\right)\text{.}
\end{align}
Using this approximated distribution, and similar to \cref{eq:gaussian_approximaged_MLE_color} and \cref{eq:gaussian_approximaged_MLE_depth}, we can obtain the approximated solutions $(\hat{\mu}_i, \hat{\sigma}_i), (\hat{\mu}_{\mathbf{c}_i}, \hat{\sigma}_{\mathbf{c}_i})$ for the volume density and color by minimizing the following negative log-likelihood:
\begin{align}
    \nonumber
    \ln \left( \sum^{N_s}_{i=1}\delta^2_i \left( \sigma^2_i\mu^2_{\mathbf{c}_i} + \sigma^2_{\mathbf{c}_i}\mu^2_i + \sigma^2_i\sigma^2_{\mathbf{c}_i} \right) \right) 
    \\
    + \frac{\left( C(\mathbf{r}) - \sum^{N_s}_{i=1}\delta_i\mu_i\mu_{\mathbf{c}_i} \right)^2}{\sum^{N_s}_{i=1} \delta^2_i \left( \sigma^2_i\mu^2_{\mathbf{c}_i} + \sigma^2_{\mathbf{c}_i}\mu^2_i + \sigma^2_i\sigma^2_{\mathbf{c}_i} \right)}. 
    \label{eq:gaussian_approximaged_MLE_color_color}
\end{align}

\subsubsection{Markov Approach for Density Uncertainty}
The approximation in \cref{eq:approximation_alpha} hardly holds when the ray $\mathbf{r}$ travels a long distance, or the volume density $\sigma$ takes a large value. As a result, the approximated solutions obtained from \cref{eq:gaussian_approximaged_MLE_color} and \cref{eq:gaussian_approximaged_MLE_color_color} become inaccurate.
To fundamentally address such approximation errors, we apply the Markov assumption to the traversal of rays as follows:
1) In \cref{eq:discrete rendering function}, $\alpha_i$ represents $i'{th}$ random variable involving $i'{th}$ density $\rho_i$ and the past densities $\rho_0,...,\rho_{i-1}$ expressed as $T_i$.
Since $\rho_0,...,\rho_{i-1}$ are variables observed along the path traveled by the ray in temporal order, they can be considered deterministic variables at time step $i$.
Therefore, the probability distribution for $\alpha_i$ can be expressed as 
$
    p\left(\alpha_i | \rho_i,..., \rho_{0}\right) 
    \simeq 
    p\left(\alpha_i | \rho_i\right)
    =
    p\left(\alpha_i | o \right).
$
Here, $o=1-\exp(-\delta_i \rho_i)$ is defined as the occupancy variable, which effectively normalizes density to indicate the presence of material.
2) Similarly, if the summation in the rendering process \cref{eq:color,eq:depth} is performed in temporal order, then $\alpha_i$ can also be regarded as a variable determined in certain time step.
In other words, at time step $i+1$, $\alpha_{0},...,\alpha_{i}$ are already obtained and determined, making them independent of $\alpha_{i+1}$.
3) We assume that occupancy $o$ follows a Gaussian distribution $o \sim \mathcal{N} \left(\mu_o, \sigma_o^2 \right)$.
Following these reasonings 1)-3), the rendered color $C\left(\mathbf{r}\right)$ in \cref{eq:color,eq:depth} can be treated as the linear combination of the Gaussian random variables, simplifying its distribution as follows:
\begin{align}
C\left(\mathbf{r}\right)
\sim
\mathcal{N}\left(
    \sum^{N_s}_{i=1}\mathbf{c}_i T_i\mu_{o_i},
    \sum^{N_s}_{i=1}\mathbf{c}^2_i T^2_i\sigma^2_{o_i}
\right)  \text{.}
\end{align}
Using the distribution of the rendered color, we can derive the approximate solution $\{\left( \hat{\mu}_i, \hat{\sigma}_i\right) \}$ for volume density by minimizing the following negative log-likelihood:
\begin{align}
    \ln \sum^{N_s}_{i=1}\mathbf{c}^2_i T^2_i \sigma^2_{o_i}
    +
    \frac
    {\left( C\left(\mathbf{r}\right) - \sum^{N_s}_{i=1}\mathbf{c}_i T_i\mu_{o_i} \right)^2}
    {\sum^{N_s}_{i=1}\mathbf{c}^2_i T^2_i \sigma^2_{o_i}}
    \text{.}
    \label{eq:markov_approximaged_MLE_color}
\end{align}

Note that for $D\left(\mathbf{r}\right)$, by replacing $\mathbf{c}_i$ to $d_i$, we can have the similar formulations for the depth estimation since we have the generalized approximation for the occupancy $o$.
We assume the probability variables to be independent with respect to the time sequence and validate the efficiency of such approximation methods through experiments.

\section{\textbf{Experiment}}
\subsection{\textbf{Experimental Setup}}
\subsubsection{Evaluation}
In our experiments, we explore the concept of unobserved and observed views in the context of NeRF. Unobserved views are perspectives not captured by the camera due to limited trajectories or self-occlusion. In contrast, observed views are near those captured by the camera. Our method estimates unobserved views without an additional network, focusing on uncertainty related to density. Additionally, we compare the predictions of observed views to analyze the performance improvements in rendering unobserved views. We evaluated five different methods: (a) \textit{Baseline}, using the vanila NeRF method with only photometric loss; (b) \textit{Color} \cite{martin2021nerf, pan2022activenerf}, integrating color uncertainty into NeRF; (c) \textit{Density}, emphasizing density uncertainty during NeRF training; 
(d) \textit{Den\_cf} \cite{shen2022conditional}, which uses a generative model to estimate density uncertainty via an additional network;
(e) \textit{Color + Density}, addressing both color and density uncertainties;
and (f) \textit{Occupancy}, advancing NeRF by considering occupancy uncertainty based on Markov assumptions.
In addition to these experiments, we extend our approach to scenarios such as SLAM, where uncertainty is prevalent. Specifically, we apply our method to NICE SLAM \cite{zhu2022nice}, which represents maps using NeRF, allowing us to validate the effectiveness of our uncertainty modeling strategy.

\begin{figure}[b]
    \centering
    \begin{subfigure}[b]{0.49\textwidth}
        \includegraphics[width=\textwidth]{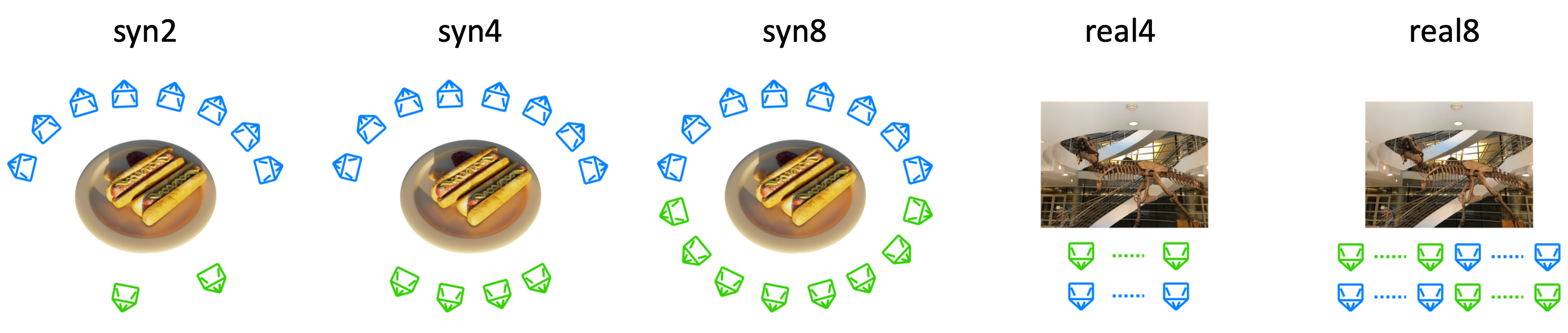}
        \caption{NeRF synthetic and real-world dataset experiment setting}
        \medskip 
    \end{subfigure}
    \begin{subfigure}[b]{0.49\textwidth}
        \includegraphics[width=\textwidth]{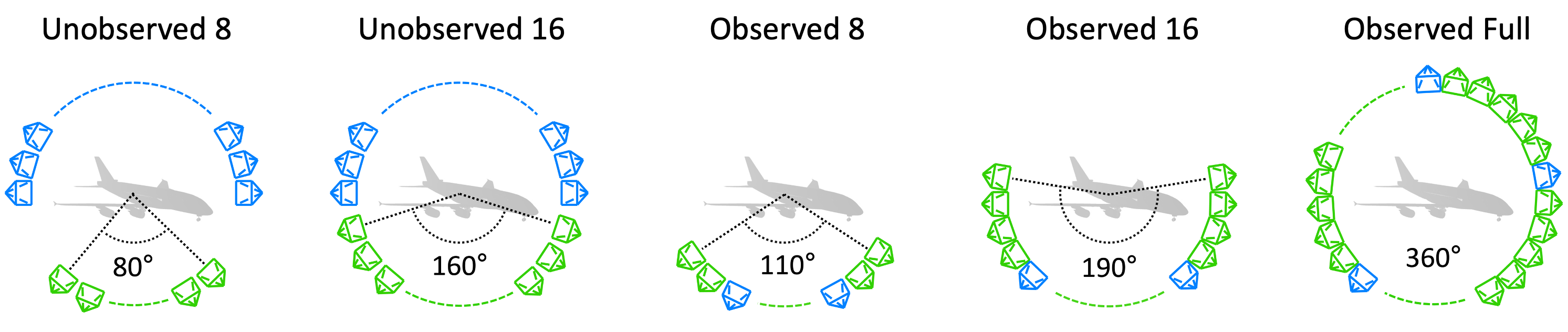}
        \caption{ModelNet dataset experiment setting}
    \end{subfigure}
    \caption{\textbf{Dataset Settings.} We conduct experiments on NeRF synthetic, real-world, and ModelNet datasets, using green cameras for training and blue cameras for testing. (a) The NeRF synthetic dataset evaluates unobserved predictions with limited training views, while the real-world dataset assesses observed predictions with forward-facing camera trajectories. (b) The ModelNet dataset evaluates both unobserved and observed predictions using 36 images spaced at 10-degree intervals.}
    \label{fig: dataset setting}
\end{figure}

\subsubsection{Metric}
In our study, we use PSNR \cite{korhonen2012peak}, SSIM \cite{yue2005similarity}, and LPIPS \cite{zhang2018unreasonable} as primary metrics for RGB image quality. For depth image evaluation, we apply Absolute Relative Error (AbsRel) and Root Mean Square Error in log space (RMSElog) to measure depth deviation with a focus on precision. We also use Average log10 Error (log10) and Threshold Accuracy ($\delta_i$) to assess depth accuracy. In the SLAM experiments, we evaluate both mapping and tracking performance. Mapping is assessed using the aforementioned metrics, along with additional 2D and 3D metrics for scene geometry. For 2D evaluation, we calculate L1 loss on depth maps by comparing reconstructed meshes to the ground truth. For 3D evaluation, we follow \cite{sucar2021imap}, focusing on Accuracy [cm], Completion [cm], and Completion Ratio [$<$ 5cm \%], excluding regions outside any camera’s viewing frustum. Mapping evaluation is conducted on images not used in keyframes, ensuring unbiased assessment. Camera tracking performance is measured using ATE RMSE.

\subsection{\textbf{Experiment Results and Discussion}}
\subsubsection{Results on NeRF Synthetic and LLFF Dataset}
\begin{table}[hbt!]
    \caption{\textbf{Quantitative results on NeRF synthetic and LLFF datasets \cite{mildenhall2021nerf}.} We compare our methods with the baseline \cite{mildenhall2021nerf}, color uncertainty methods Col \cite{martin2021nerf, pan2022activenerf}, and Den\_cf, which estimates density uncertainty using a generative model \cite{shen2022conditional}. Experiments are conducted across all scenes, and average values are reported. 
    Significant improvements are observed in both unobserved and observed views, with methods addressing occupancy and combined color-density uncertainties showing superior performance. "Full" indicates using the entire training dataset.}
    \label{tab:NeRF_dataset}
    \centering

    \begin{tabular}{c|cccccc}
        \multicolumn{7}{c}{PSNR $\uparrow$} \\
        \;Dataset\; & \;Base \; & \;Col \; & \;Den\; & \;Den\_cf\; & \;Col+Den\; & \;Occu\; \\
        \hline
        Syn2 & 11.75 & 11.29 & 13.61 & 12.8 & 14.72 & \textbf{15.12} \\
        Syn4 & 15.40 & 16.18 & 17.02 & 15.9 & 18.01 & \textbf{18.30} \\
        Syn8 & 18.04 & 18.91 & 19.19 & 18.5 & \textbf{20.87} & 20.75 \\
        SynF & 25.57 & 24.20 & 24.25 & 21.2 & 25.64 & \textbf{25.68} \\
        \hline
        Real4 & 12.88 & 12.97 & 12.88 & 14.0 & \textbf{13.51} & 13.45 \\
        Real8 & 22.60 & 22.66 & 22.26 & 20.3 & 23.01 & \textbf{23.10} \\
        RealF & 24.81 & 24.03 & 23.86 & 21.4 & 24.78 & \textbf{24.97} \\
        \hline
    \end{tabular}
    \medskip 
    
    \begin{tabular}{c|cccccc}
        \multicolumn{7}{c}{SSIM $\uparrow$} \\
        \;Dataset\; & \;Base \; & \;Col \; & \;Den\; & \;Den\_cf\; & \;Col+Den\; & \;Occu\; \\
        \hline
        Syn2 & 0.61 & 0.61 & 0.66 & \textbf{0.71} & 0.67 & 0.69 \\
        Syn4 & 0.74 & 0.75 & 0.76 & \textbf{0.79} & 0.78 & \textbf{0.79} \\
        Syn8 & 0.79 & 0.80 & 0.79 & 0.80 & \textbf{0.82} & \textbf{0.82} \\
        SynF & \textbf{0.88} & 0.86 & 0.86 & 0.82 & \textbf{0.88} & \textbf{0.88} \\
        \hline
        Real4 & 0.52 & 0.53 & 0.52 & \textbf{0.58} & 0.54 & 0.55 \\
        Real8 & 0.69 & 0.69 & 0.67 & 0.66 & 0.69 & \textbf{0.71} \\
        RealF & 0.76 & 0.75 & 0.73 & 0.70 & 0.76 & \textbf{0.77} \\
        \hline
    \end{tabular}
    \medskip 
    
    \begin{tabular}{c|cccccc}
        \multicolumn{7}{c}{LPIPS $\downarrow$} \\
        \;Dataset\; & \;Base \; & \;Col \; & \;Den\; & \;Den\_cf\; & \;Col+Den\; & \;Occu\; \\
        \hline
        Syn2 & 0.44 & 0.42 & 0.36 & 0.39 & 0.34 & \textbf{0.33} \\
        Syn4 & 0.27 & 0.26 & 0.24 & 0.29 & 0.21 & \textbf{0.20} \\
        Syn8 & 0.22 & 0.21 & 0.21 & 0.24 & \textbf{0.17} & \textbf{0.17} \\
        SynF & 0.11 & 0.14 & 0.14 & 0.22 & 0.11 & \textbf{0.10} \\
        \hline
        Real4 & 0.62 & 0.63 & 0.62 & 0.63 & 0.60 & \textbf{0.59} \\
        Real8 & 0.22 & 0.24 & 0.25 & 0.47 & 0.20 & \textbf{0.19} \\
        RealF & 0.17 & 0.21 & 0.22 & 0.42 & 0.17 & \textbf{0.16} \\
        \hline
    \end{tabular}
\end{table}
\begin{figure}[t]
  \centering
  \includegraphics[width=0.45\textwidth]{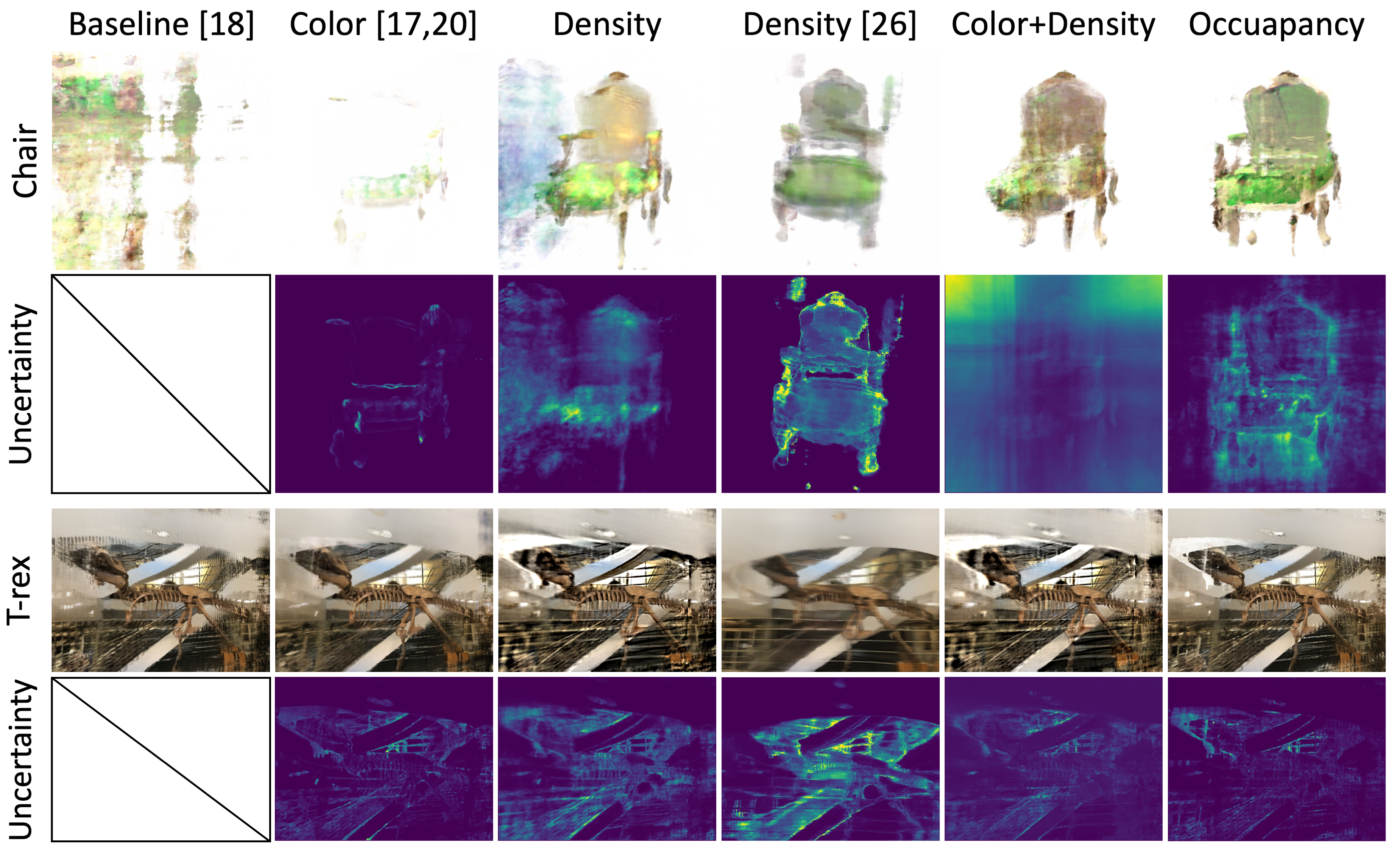}
  \caption{
  \textbf{Qualitative Comparisons on NeRF synthetic and LLFF datasets \cite{mildenhall2021nerf}.} 
  The Chair scene is from the NeRF synthetic dataset trained with the syn4 setting, and the T-rex scene is from the LLFF dataset trained with the real8 setting. These settings are described in \cref{fig: dataset setting}. While incorporating uncertainty improves performance, the (b) \textit{Color} method in synthetic datasets struggles with density estimation, causing objects to appear faded.
}
  \label{fig:res1}
  \vspace{-5pt}
\end{figure}
In \cref{tab:NeRF_dataset}, we show the RGB image metric results for the six methods. The synthetic dataset shows significant improvements, while the real-world dataset shows modest gains for forward-facing images. Additionally, settings with fewer training images yield greater performance improvements than those using the full number of images in the dataset. This outcome indicates that in scenarios where data is limited, the inherent uncertainty within the data becomes more pronounced, making the integration of uncertainty into the learning process significantly more impactful.

While considering uncertainty generally improves performance, the (b) \textit{Color} and (c) \textit{Density} approaches show relatively low performance, especially with fewer training images. The (b) \textit{Color} method struggles to estimate density accurately under these conditions, leading to visual artifacts, such as objects appearing faded or vanishing. This limitation highlights the (b) \textit{Color} method's inability to capture detailed density variations with limited data. In contrast, methods that account for uncertainty in density demonstrate improved robustness. These methods preserve objects' visibility and structural integrity, emphasizing the critical role of managing uncertainty in geometric structures.

However, even methods that consider density have their own issues. In our experiments, the (c) \textit{Density} method revealed that high-density datasets lead to challenges for accurate estimation. This results in a decline in performance and a less reliable MLE formulation, with the (c) \textit{Density} method underperforming compared to the (e) \textit{Color + Density} and (f) \textit{Occupancy} approaches. Additionally, the (d) \textit{Den\_cf} method, which employs a generative model to estimate density uncertainty via an additional network, shows inferior performance in PSNR and LPIPS metrics, especially on synthetic datasets with fewer training images. This occurs because neural networks simplify complex probability distributions, hindering the capture of fine textures and details. However, it performs well in the SSIM metric by fitting the complex distribution as closely as possible, effectively capturing the overall structure of the object despite not precisely modeling the probability distribution.
These issues are effectively addressed by the (e) \textit{Color + Density} approach, which integrates both methods and shows significant performance improvements. However, it occasionally fails to meet the fundamental assumptions of the (c) \textit{Density} approach, leading to suboptimal performance when volume density assumptions are not fully satisfied. In contrast, the (f) \textit{Occupancy} approach, based on reasonable Markov assumptions, exhibits remarkable stability across both synthetic and real-world datasets, managing inherent uncertainty without relying heavily on specific assumptions.

\subsubsection{Results on ModelNet Dataset}
\begin{table*}[t!]
\centering
    \caption{\textbf{Quantitative results on ModelNet \cite{wu20153d}.} Given that depth images lack color information, we compare the (a) \textit{Baseline}, (c) \textit{Density}, and (e) \textit{Occupancy} methods. Experiments were conducted on two instances across all classes, with average values provided. The dataset setting follows \cref{fig: dataset setting}, and similar to the RGB dataset, considering uncertainty in the unobserved view prediction setting demonstrates significant performance improvements.}
    \label{tab:modelnet result}
    \centering

    \begin{tabular}{c|ccc|ccc}
        \multicolumn{7}{c}{Unobserved View Setting} \\
        \;Dataset\;      & \;$\delta < 1.25\uparrow$\; & \;$\delta < 1.25^2\uparrow$\;  & \;$\delta < 1.25^3\uparrow$\; & \;AbsRel$\downarrow$\;   & \;RMSE(log)$\downarrow$\;  & \;log10$\downarrow$\;   \\
        \hline
        Baseline 8 \cite{mildenhall2021nerf}    & 0.497 & 0.672 & 0.755 & 0.537 & 0.467 & 0.159 \\ 
        Uncert den 8  & 0.506 & 0.685 & 0.809 & 0.492 & 0.440 & 0.151 \\
        Uncert den\_cf 8 \cite{shen2022conditional}  & \textbf{0.568} & 0.709 & \textbf{0.841} & \textbf{0.448} & \textbf{0.403} & \textbf{0.137} \\
        Uncert occu 8  & 0.543 & \textbf{0.712} & 0.837 & 0.451 & 0.418 & 0.140 \\
        \hline
        Baseline 16 \cite{mildenhall2021nerf}  & 0.740 & 0.906 & 0.962 & 0.203 & 0.262 & 0.076 \\
        Uncert den 16 & 0.762 & 0.911 & 0.961 & 0.192 & 0.257 & 0.073 \\ 
        Uncert den\_cf 16 \cite{shen2022conditional}  & \textbf{0.788} & 0.914 & \textbf{0.968} & 0.182 & 0.249 & 0.070 \\
        Uncert occu 16 & 0.785 & \textbf{0.922} & 0.964 & \textbf{0.177} & \textbf{0.245} & \textbf{0.068} \\
        \hline
    \end{tabular}
    \medskip
    
    \begin{tabular}{c|ccc|ccc}
        \multicolumn{7}{c}{Observed View Setting} \\
        \;Dataset\;      & \;$\delta < 1.25\uparrow$\; & \;$\delta < 1.25^2\uparrow$\;  & \;$\delta < 1.25^3\uparrow$\; & \;AbsRel$\downarrow$\;   & \;RMSE(log)$\downarrow$\;  & \;log10$\downarrow$\;   \\
        \hline
        Baseline 8 \cite{mildenhall2021nerf}   & 0.880 & 0.939 & 0.971 & 0.102 & 0.192 & 0.044 \\ 
        Uncert den 8  & 0.886 & 0.943 & 0.973 & 0.099 & 0.188 & 0.043 \\
        Uncert den\_cf 8 \cite{shen2022conditional}  & \textbf{0.894} & 0.945 & \textbf{0.978} & \textbf{0.095} & \textbf{0.183} & \textbf{0.039} \\
        Uncert occu 8  & 0.891 & \textbf{0.946} & 0.975 & 0.096 & 0.184 & 0.041 \\ 
        \hline
        Baseline 16 \cite{mildenhall2021nerf}  & 0.888 & 0.943 & 0.974 & 0.099 & 0.189 & 0.043 \\
        Uncert den 16 & 0.891 & 0.946 & 0.977 & 0.097 & 0.186 & 0.042 \\ 
        Uncert den\_cf 16 \cite{shen2022conditional}  & 0.892 & \textbf{0.952} & 0.977 & \textbf{0.094} & \textbf{0.182} & \textbf{0.041} \\
        Uncert occu 16 & \textbf{0.895} & 0.949 & \textbf{0.978} & \textbf{0.094} & 0.183 & \textbf{0.041} \\      
        \hline
        Baseline Full \cite{mildenhall2021nerf}  & 0.896 & 0.946 & 0.976 & 0.096 & 0.180 & 0.041 \\
        Uncert den Full & 0.896 & 0.946 & 0.977 & 0.096 & 0.181 & 0.041 \\ 
        Uncert den\_cf Full \cite{shen2022conditional}  & 0.898 & 0.945 & 0.976 & 0.095 & 0.178 & 0.040 \\
        Uncert occu Full & \textbf{0.902} & \textbf{0.949} & \textbf{0.978} & \textbf{0.092} & \textbf{0.176} & \textbf{0.039} \\
        \hline
    \end{tabular}
\end{table*}

\begin{figure}[hbt!]
  \centering
  \includegraphics[width=0.44\textwidth]{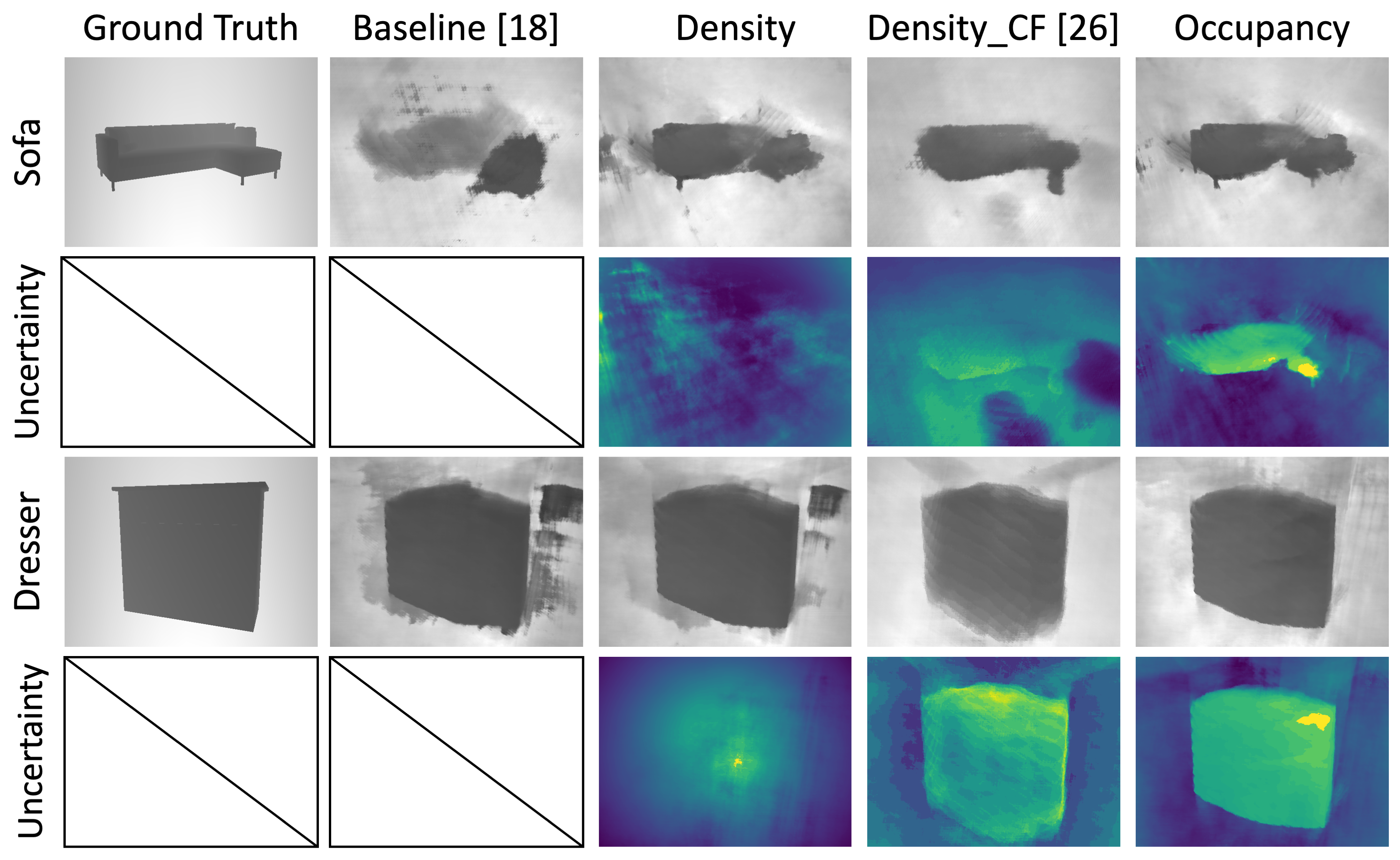}
  \caption{
  \textbf{Qualitative Comparisons on ModelNet \cite{wu20153d}}.
  The sofa and dresser scene employs an unobserved view setting and is trained using 8 images. In this scene, the (d) \textit{Den\_cf} and (f) \textit{Occupancy} methods effectively addresses the blurring issue.
  }
  \label{fig:res2}
  \vspace{-5pt}
\end{figure}
In our ModelNet dataset analysis, as shown in \cref{tab:modelnet result}, we present the metric results for depth images, where the absence of RGB information requires focusing on geometric reconstruction. We compare the (a) \textit{Baseline}, (c) \textit{Density}, (d) \textit{Den\_cf}, and (f) \textit{Occupancy} methods. Our findings align with the trends seen in RGB images, showing significant improvements in unobserved views, while observed views exhibit more modest gains.
However, unlike the results in RGB images, in the unobserved view setting with limited training images, the (d) \textit{Den\_cf} method shows reasonably good performance. This is because depth image prediction does not require the estimation of textures or fine details; instead, the primary goal is to predict the object's shape from a limited number of training images. In such cases, approximating the complex distribution as accurately as possible using a generative model proves to be effective.
For instance, in the sofa and dresser scene from \cref{fig:res2}, blurring in empty spaces is significantly reduced by incorporating uncertainty, particularly with the (d) \textit{Den\_cf} and (f) \textit{Occupancy} methods, which refine the clarity of the depth images more effectively than the (c) \textit{Density} method.
In settings where all training images are used, the importance shifts to predicting detailed aspects beyond just the object's shape. 
Here, our (f) \textit{Occupancy} method, which explicitly estimates occupancy uncertainty, demonstrates superior performance.

\subsubsection{Results in SLAM Environment}
\begin{table*}[h]
\centering
\caption{\textbf{Quantitative results on Replica \cite{straub2019replica} and TUM \cite{sturm2012benchmark}.} We compare the mapping and tracking performance of the uncertainty method with the original method. The results are averaged across the room and office scenes in the Replica dataset and three scenes in the TUM dataset. Scenarios are denoted as follows: \_1 for all data used, \_2 for even-indexed images only, and \_3 for images indexed by multiples of 3.}
\label{tab:slam_table}
\begin{tabular}{l|l|c c c|c c c|c c c}
\toprule
Methods & Metrics & RAvg$_1$ & OAvg$_1$ & TAvg$_1$ & RAvg$_2$ & OAvg$_2$ & TAvg$_2$ & RAvg$_3$ & OAvg$_3$ & TAvg$_3$ \\
\midrule
\multirow{8}{*}{Baseline Method \cite{zhu2022nice}}
& PSNR $\uparrow$ & 22.48 & 24.49 & 14.66 & 17.35 & 20.31 & 12.64 & 13.82 & 15.86 & 13.21 \\
& SSIM $\uparrow$ & 0.75 & 0.82 & 0.47 & 0.57 & 0.68 & 0.35 & 0.43 & \textbf{0.56} & \textbf{0.38} \\
& LPIPS $\downarrow$ & 0.43 & 0.34 & \textbf{0.54} & 0.56 & 0.54 & 0.61 & 0.66 & 0.56 & 0.60 \\
& RMSE $\downarrow$ & \textbf{0.05} & 0.05 & \textbf{0.04} & 0.49 & 0.68 & 0.21 & 1.08 & 0.87 & 0.20 \\
& Acc $\downarrow$ & \textbf{3.12} & 4.20 & \text{--} & 32.1 & 17.7 & \text{--} & 59.6 & 40.8 & \text{--} \\
& Comp $\downarrow$ & \textbf{3.16} & 4.01 & \text{--} & 13.6 & 9.44 & \text{--} & 36.4 & 19.6 & \text{--} \\
& Comp Ratio $\uparrow$ & \textbf{87.9} & 81.2 & \text{--} & 38.5 & 56.3 & \text{--} & 18.8 & 39.0 & \text{--} \\
& Depth L1 $\downarrow$ & \textbf{8.12} & 24.3 & \text{--} & 104.5 & 75.6 & \text{--} & 136.3 & 150.6 & \text{--} \\
\midrule
\multirow{8}{*}{Uncertainty Method}
& PSNR $\uparrow$ & \textbf{22.74} & \textbf{24.94} & \textbf{15.05} & \textbf{18.82} & \textbf{20.45} & \textbf{13.67} & \textbf{15.25} & \textbf{16.63} & \textbf{13.33} \\
& SSIM $\uparrow$ & \textbf{0.77} & \textbf{0.84} & \textbf{0.49} & \textbf{0.62} & \textbf{0.68} & \textbf{0.43} & \textbf{0.49} & 0.55 & 0.37 \\
& LPIPS $\downarrow$ & \textbf{0.42} & \textbf{0.33} & 0.55 & \textbf{0.53} & \textbf{0.44} & \textbf{0.47} & \textbf{0.62} & \textbf{0.55} & \textbf{0.59} \\
& RMSE $\downarrow$ & 0.06 & \textbf{0.04} & 0.06 & \textbf{0.43} & \textbf{0.28} & \textbf{0.10} & \textbf{0.78} & \textbf{0.73} & \textbf{0.17} \\
& Acc $\downarrow$ & 3.71 & \textbf{3.28} & \text{--} & \textbf{20.7} & \textbf{13.3} & \text{--} & \textbf{42.5} & \textbf{33.3} & \text{--} \\
& Comp $\downarrow$ & 4.29 & \textbf{3.87} & \text{--} & \textbf{12.7} & \textbf{8.87} & \text{--} & \textbf{35.9} & \textbf{17.6} & \text{--} \\
& Comp Ratio $\uparrow$ & 79.7 & \textbf{83.1} & \text{--} & \textbf{49.7} & \textbf{57.4} & \text{--} & \textbf{19.5} & \textbf{40.8} & \text{--} \\
& Depth L1 $\downarrow$ & 12.2 & \textbf{13.6} & \text{--} & \textbf{57.8} & \textbf{74.3} & \text{--} & \textbf{132.3} & \textbf{137.2} & \text{--} \\
\bottomrule
\end{tabular}
\end{table*}

\begin{figure}[bt!]
  \centering
  \includegraphics[width=0.45\textwidth]{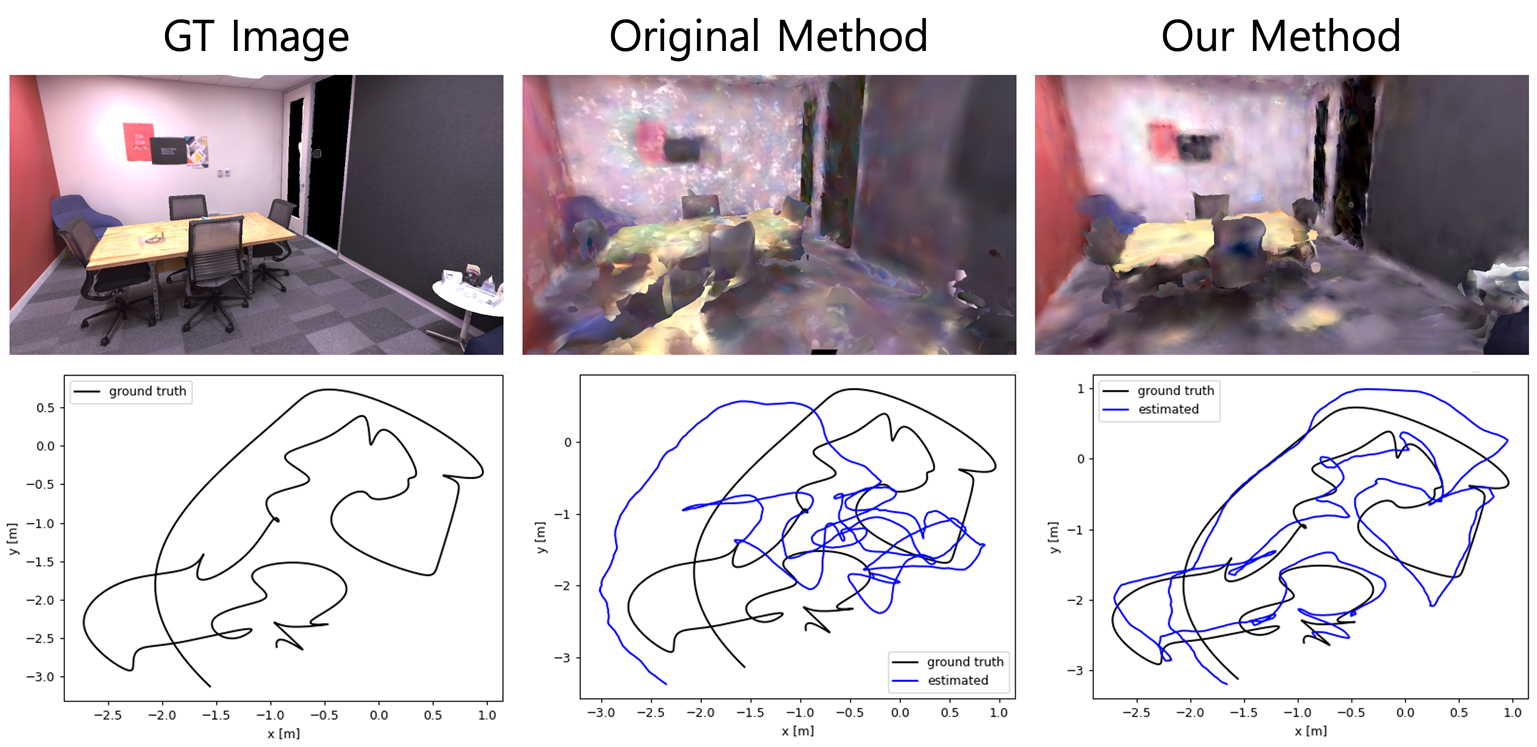}
  \caption{\textbf{Mapping and Tracking Visualization Results on the Replica Office2 scene.} 
  The figure shows intermediate rendering results during the mapping process, which are not used for training. Our method demonstrates improvements over the original method in both mapping and tracking, even with limited data.}
  \label{fig:slam_result}
  \vspace{-5pt}
\end{figure}

In our previous experiments, we demonstrated the effectiveness of training neural fields by incorporating uncertainty, particularly by adjusting the level of uncertainty inherent in the data. Building on these results, we further validate our approach in NICE SLAM \cite{zhu2022nice}, a scenario with varying levels of uncertainty. 
However, existing methods that estimate uncertainty using generative models, such as CF-NeRF \cite{shen2022conditional}, require significantly more computational resources. Under the vanilla NeRF settings, we observed that CF-NeRF's training time is approximately 21 times longer, and the runtime per image is about 11 times longer than the baseline NeRF. These substantial computational demands make such methods unsuitable for real-time applications like SLAM, where both training and rendering need to be performed efficiently. 
In contrast, our approach explicitly estimates uncertainty without relying on additional networks, resulting in training and runtime that are only approximately 1.01 times and 1.04 times longer than the baseline. Therefore, we choose to focus on occupancy uncertainty, as it is suitable for real-time applications and has demonstrated good performance in previous experiments.

As demonstrated in \cref{tab:slam_table}, when using all available images for training, the performance difference between methods that consider uncertainty and those that do negligible, since comprehensive data coverage minimizes spatial uncertainty.
However, when only every second or third image is used, thereby increasing spatial uncertainty, our method that incorporates uncertainty significantly outperforms the baseline. To further test the robustness of our approach, we conducted additional experiments where we adjusted key hyperparameters, such as the mapping frequency and keyframe selection interval, to compensate for the reduced data. Even with these adjustments, our uncertainty method continued to outperform the baseline, demonstrating its effectiveness in challenging environments with limited data.
This improved performance is visually evident in \cref{fig:slam_result}, which illustrates intermediate rendering results during the mapping process. The original method without uncertainty produces less accurate scene representations under limited data conditions. In contrast, our method provides more robust and accurate renderings, highlighting the benefits of incorporating uncertainty. Additionally, our estimated trajectory aligns more closely with the ground truth, enhancing tracking performance. These results emphasize that considering uncertainty improves both mapping and tracking, especially in limited data scenarios.

\section{\textbf{CONCLUSIONS}}

In this work, we introduced a Bayesian Neural Radiance Field (NeRF) method, significantly advancing 3D scene representation from various viewpoints. Our approach effectively quantifies uncertainty in geometric volume structures without relying on additional networks, making it robust in handling challenging observations. By implementing generalized approximations and defining density-related uncertainty, we have extended the application of Bayesian NeRF beyond RGB images to depth images, achieving enhanced performance across extensive datasets. Furthermore, we validated our method in a SLAM environment, demonstrating that incorporating occupancy uncertainty enhances both mapping and tracking performance. Overall, our proposed extensions to NeRF offer a scalable and efficient solution for managing uncertainty in complex environments, contributing to improved performance in real-world applications with limited data availability.



{
\footnotesize
\bibliographystyle{IEEEtranS}
\bibliography{main}

\begin{thebibliography}{10}
\providecommand{\url}[1]{#1}
\csname url@rmstyle\endcsname
\providecommand{\newblock}{\relax}
\providecommand{\bibinfo}[2]{#2}
\providecommand\BIBentrySTDinterwordspacing{\spaceskip=0pt\relax}
\providecommand\BIBentryALTinterwordstretchfactor{4}
\providecommand\BIBentryALTinterwordspacing{\spaceskip=\fontdimen2\font plus
\BIBentryALTinterwordstretchfactor\fontdimen3\font minus \fontdimen4\font\relax}
\providecommand\BIBforeignlanguage[2]{{%
\expandafter\ifx\csname l@#1\endcsname\relax
\typeout{** WARNING: IEEEtran.bst: No hyphenation pattern has been}%
\typeout{** loaded for the language `#1'. Using the pattern for}%
\typeout{** the default language instead.}%
\else
\language=\csname l@#1\endcsname
\fi
#2}}

\bibitem{abdar2021review}
M.~Abdar, F.~Pourpanah, S.~Hussain, D.~Rezazadegan, L.~Liu, M.~Ghavamzadeh, P.~Fieguth, X.~Cao, A.~Khosravi, U.~R. Acharya, \emph{et~al.}, ``A review of uncertainty quantification in deep learning: Techniques, applications and challenges,'' \emph{Information fusion}, vol.~76, pp. 243--297, 2021.

\bibitem{barron2023zip}
J.~T. Barron, B.~Mildenhall, D.~Verbin, P.~P. Srinivasan, and P.~Hedman, ``Zip-nerf: Anti-aliased grid-based neural radiance fields,'' \emph{arXiv preprint arXiv:2304.06706}, 2023.

\bibitem{blundell2015weight}
C.~Blundell, J.~Cornebise, K.~Kavukcuoglu, and D.~Wierstra, ``Weight uncertainty in neural network,'' in \emph{International conference on machine learning}.\hskip 1em plus 0.5em minus 0.4em\relax PMLR, 2015, pp. 1613--1622.

\bibitem{deng2022fov}
N.~Deng, Z.~He, J.~Ye, B.~Duinkharjav, P.~Chakravarthula, X.~Yang, and Q.~Sun, ``Fov-nerf: Foveated neural radiance fields for virtual reality,'' \emph{IEEE Transactions on Visualization and Computer Graphics}, vol.~28, no.~11, pp. 3854--3864, 2022.

\bibitem{goli2023bayes}
L.~Goli, C.~Reading, S.~Selll{\'a}n, A.~Jacobson, and A.~Tagliasacchi, ``Bayes' rays: Uncertainty quantification for neural radiance fields,'' \emph{arXiv preprint arXiv:2309.03185}, 2023.

\bibitem{hu2023tri}
W.~Hu, Y.~Wang, L.~Ma, B.~Yang, L.~Gao, X.~Liu, and Y.~Ma, ``Tri-miprf: Tri-mip representation for efficient anti-aliasing neural radiance fields,'' in \emph{Proceedings of the IEEE/CVF International Conference on Computer Vision}, 2023, pp. 19\,774--19\,783.

\bibitem{hu2023pc}
X.~Hu, G.~Xiong, Z.~Zang, P.~Jia, Y.~Han, and J.~Ma, ``Pc-nerf: Parent-child neural radiance fields under partial sensor data loss in autonomous driving environments,'' \emph{arXiv preprint arXiv:2310.00874}, 2023.

\bibitem{huang2023neural}
S.~Huang, Z.~Gojcic, Z.~Wang, F.~Williams, Y.~Kasten, S.~Fidler, K.~Schindler, and O.~Litany, ``Neural lidar fields for novel view synthesis,'' \emph{arXiv preprint arXiv:2305.01643}, 2023.

\bibitem{kendall2017uncertainties}
A.~Kendall and Y.~Gal, ``What uncertainties do we need in bayesian deep learning for computer vision?'' \emph{Advances in neural information processing systems}, vol.~30, 2017.

\bibitem{kong2023vmap}
X.~Kong, S.~Liu, M.~Taher, and A.~J. Davison, ``vmap: Vectorised object mapping for neural field slam,'' in \emph{Proceedings of the IEEE/CVF Conference on Computer Vision and Pattern Recognition}, 2023, pp. 952--961.

\bibitem{korhonen2012peak}
J.~Korhonen and J.~You, ``Peak signal-to-noise ratio revisited: Is simple beautiful?'' in \emph{2012 Fourth International Workshop on Quality of Multimedia Experience}.\hskip 1em plus 0.5em minus 0.4em\relax IEEE, 2012, pp. 37--38.

\bibitem{lee2023just}
M.~Lee, K.~Kang, and H.~Yu, ``Just flip: Flipped observation generation and optimization for neural radiance fields to cover unobserved view,'' \emph{arXiv preprint arXiv:2303.06335}, 2023.

\bibitem{liu2023zero}
R.~Liu, R.~Wu, B.~Van~Hoorick, P.~Tokmakov, S.~Zakharov, and C.~Vondrick, ``Zero-1-to-3: Zero-shot one image to 3d object,'' in \emph{Proceedings of the IEEE/CVF International Conference on Computer Vision}, 2023, pp. 9298--9309.

\bibitem{liu2023visualization}
Y.~Liu, X.~Tu, D.~Chen, K.~Han, O.~Altintas, H.~Wang, and J.~Xie, ``Visualization of mobility digital twin: Framework design, case study, and future challenges,'' in \emph{2023 IEEE 20th International Conference on Mobile Ad Hoc and Smart Systems (MASS)}.\hskip 1em plus 0.5em minus 0.4em\relax IEEE, 2023, pp. 170--177.

\bibitem{lo2012sum}
C.-F. Lo \emph{et~al.}, ``The sum and difference of two lognormal random variables,'' \emph{Journal of Applied Mathematics}, vol. 2012, 2012.

\bibitem{malik2023transient}
A.~Malik, P.~Mirdehghan, S.~Nousias, K.~N. Kutulakos, and D.~B. Lindell, ``Transient neural radiance fields for lidar view synthesis and 3d reconstruction,'' \emph{arXiv preprint arXiv:2307.09555}, 2023.

\bibitem{martin2021nerf}
R.~Martin-Brualla, N.~Radwan, M.~S. Sajjadi, J.~T. Barron, A.~Dosovitskiy, and D.~Duckworth, ``Nerf in the wild: Neural radiance fields for unconstrained photo collections,'' in \emph{Proceedings of the IEEE/CVF Conference on Computer Vision and Pattern Recognition}, 2021, pp. 7210--7219.

\bibitem{mildenhall2021nerf}
B.~Mildenhall, P.~P. Srinivasan, M.~Tancik, J.~T. Barron, R.~Ramamoorthi, and R.~Ng, ``Nerf: Representing scenes as neural radiance fields for view synthesis,'' \emph{Communications of the ACM}, vol.~65, no.~1, pp. 99--106, 2021.

\bibitem{muller2022instant}
T.~M{\"u}ller, A.~Evans, C.~Schied, and A.~Keller, ``Instant neural graphics primitives with a multiresolution hash encoding,'' \emph{ACM Transactions on Graphics (ToG)}, vol.~41, no.~4, pp. 1--15, 2022.

\bibitem{pan2022activenerf}
X.~Pan, Z.~Lai, S.~Song, and G.~Huang, ``Activenerf: Learning where to see with uncertainty estimation,'' in \emph{European Conference on Computer Vision}.\hskip 1em plus 0.5em minus 0.4em\relax Springer, 2022, pp. 230--246.

\bibitem{poole2022dreamfusion}
B.~Poole, A.~Jain, J.~T. Barron, and B.~Mildenhall, ``Dreamfusion: Text-to-3d using 2d diffusion,'' \emph{arXiv preprint arXiv:2209.14988}, 2022.

\bibitem{rosinol2023nerf}
A.~Rosinol, J.~J. Leonard, and L.~Carlone, ``Nerf-slam: Real-time dense monocular slam with neural radiance fields,'' in \emph{2023 IEEE/RSJ International Conference on Intelligent Robots and Systems (IROS)}.\hskip 1em plus 0.5em minus 0.4em\relax IEEE, 2023, pp. 3437--3444.

\bibitem{schonberger2016structure}
J.~L. Schonberger and J.-M. Frahm, ``Structure-from-motion revisited,'' in \emph{Proceedings of the IEEE conference on computer vision and pattern recognition}, 2016, pp. 4104--4113.

\bibitem{seijas2012approach}
A.~Seijas-Mac{\'\i}as and A.~Oliveira, ``An approach to distribution of the product of two normal variables,'' \emph{Discussiones Mathematicae Probability and Statistics}, vol.~32, no. 1-2, pp. 87--99, 2012.

\bibitem{seitz2006comparison}
S.~M. Seitz, B.~Curless, J.~Diebel, D.~Scharstein, and R.~Szeliski, ``A comparison and evaluation of multi-view stereo reconstruction algorithms,'' in \emph{2006 IEEE computer society conference on computer vision and pattern recognition (CVPR'06)}, vol.~1.\hskip 1em plus 0.5em minus 0.4em\relax IEEE, 2006, pp. 519--528.

\bibitem{shen2022conditional}
J.~Shen, A.~Agudo, F.~Moreno-Noguer, and A.~Ruiz, ``Conditional-flow nerf: Accurate 3d modelling with reliable uncertainty quantification,'' in \emph{European Conference on Computer Vision}.\hskip 1em plus 0.5em minus 0.4em\relax Springer, 2022, pp. 540--557.

\bibitem{shen2021stochastic}
J.~Shen, A.~Ruiz, A.~Agudo, and F.~Moreno-Noguer, ``Stochastic neural radiance fields: Quantifying uncertainty in implicit 3d representations,'' in \emph{2021 International Conference on 3D Vision (3DV)}.\hskip 1em plus 0.5em minus 0.4em\relax IEEE, 2021, pp. 972--981.

\bibitem{straub2019replica}
J.~Straub, T.~Whelan, L.~Ma, Y.~Chen, E.~Wijmans, S.~Green, J.~J. Engel, R.~Mur-Artal, C.~Ren, S.~Verma, \emph{et~al.}, ``The replica dataset: A digital replica of indoor spaces,'' \emph{arXiv preprint arXiv:1906.05797}, 2019.

\bibitem{sturm2012benchmark}
J.~Sturm, N.~Engelhard, F.~Endres, W.~Burgard, and D.~Cremers, ``A benchmark for the evaluation of rgb-d slam systems,'' in \emph{2012 IEEE/RSJ international conference on intelligent robots and systems}.\hskip 1em plus 0.5em minus 0.4em\relax IEEE, 2012, pp. 573--580.

\bibitem{sucar2021imap}
E.~Sucar, S.~Liu, J.~Ortiz, and A.~J. Davison, ``imap: Implicit mapping and positioning in real-time,'' in \emph{Proceedings of the IEEE/CVF international conference on computer vision}, 2021, pp. 6229--6238.

\bibitem{sunderhauf2023density}
N.~S{\"u}nderhauf, J.~Abou-Chakra, and D.~Miller, ``Density-aware nerf ensembles: Quantifying predictive uncertainty in neural radiance fields,'' in \emph{2023 IEEE International Conference on Robotics and Automation (ICRA)}.\hskip 1em plus 0.5em minus 0.4em\relax IEEE, 2023, pp. 9370--9376.

\bibitem{tancik2022block}
M.~Tancik, V.~Casser, X.~Yan, S.~Pradhan, B.~Mildenhall, P.~P. Srinivasan, J.~T. Barron, and H.~Kretzschmar, ``Block-nerf: Scalable large scene neural view synthesis,'' in \emph{Proceedings of the IEEE/CVF Conference on Computer Vision and Pattern Recognition}, 2022, pp. 8248--8258.

\bibitem{tao2023lidar}
T.~Tao, L.~Gao, G.~Wang, P.~Chen, D.~Hao, X.~Liang, M.~Salzmann, and K.~Yu, ``Lidar-nerf: Novel lidar view synthesis via neural radiance fields,'' \emph{arXiv preprint arXiv:2304.10406}, 2023.

\bibitem{thrun2002probabilistic}
S.~Thrun, ``Probabilistic robotics,'' \emph{Communications of the ACM}, vol.~45, no.~3, pp. 52--57, 2002.

\bibitem{Turki_2022_CVPR}
H.~Turki, D.~Ramanan, and M.~Satyanarayanan, ``Mega-nerf: Scalable construction of large-scale nerfs for virtual fly-throughs,'' in \emph{Proceedings of the IEEE/CVF Conference on Computer Vision and Pattern Recognition (CVPR)}, June 2022, pp. 12\,922--12\,931.

\bibitem{wang2022clip}
C.~Wang, M.~Chai, M.~He, D.~Chen, and J.~Liao, ``Clip-nerf: Text-and-image driven manipulation of neural radiance fields,'' in \emph{Proceedings of the IEEE/CVF Conference on Computer Vision and Pattern Recognition}, 2022, pp. 3835--3844.

\bibitem{wang2023co}
H.~Wang, J.~Wang, and L.~Agapito, ``Co-slam: Joint coordinate and sparse parametric encodings for neural real-time slam,'' in \emph{Proceedings of the IEEE/CVF Conference on Computer Vision and Pattern Recognition}, 2023, pp. 13\,293--13\,302.

\bibitem{wu20153d}
Z.~Wu, S.~Song, A.~Khosla, F.~Yu, L.~Zhang, X.~Tang, and J.~Xiao, ``3d shapenets: A deep representation for volumetric shapes,'' in \emph{Proceedings of the IEEE conference on computer vision and pattern recognition}, 2015, pp. 1912--1920.

\bibitem{wu2023mars}
Z.~Wu, T.~Liu, L.~Luo, Z.~Zhong, J.~Chen, H.~Xiao, C.~Hou, H.~Lou, Y.~Chen, R.~Yang, \emph{et~al.}, ``Mars: An instance-aware, modular and realistic simulator for autonomous driving,'' in \emph{CAAI International Conference on Artificial Intelligence}.\hskip 1em plus 0.5em minus 0.4em\relax Springer, 2023, pp. 3--15.

\bibitem{lognorm}
T.~Wutzler, ``{Functions for the lognormal distribution in R}, howpublished = {\url{https://github.com/bgctw/lognorm}}, note = {Accessed: 2023-09-19}.''

\bibitem{xiang2015data}
Y.~Xiang, W.~Choi, Y.~Lin, and S.~Savarese, ``Data-driven 3d voxel patterns for object category recognition,'' in \emph{Proceedings of the IEEE conference on computer vision and pattern recognition}, 2015, pp. 1903--1911.

\bibitem{yang2023freenerf}
J.~Yang, M.~Pavone, and Y.~Wang, ``Freenerf: Improving few-shot neural rendering with free frequency regularization,'' in \emph{Proceedings of the IEEE/CVF Conference on Computer Vision and Pattern Recognition}, 2023, pp. 8254--8263.

\bibitem{yang2022vox}
X.~Yang, H.~Li, H.~Zhai, Y.~Ming, Y.~Liu, and G.~Zhang, ``Vox-fusion: Dense tracking and mapping with voxel-based neural implicit representation,'' in \emph{2022 IEEE International Symposium on Mixed and Augmented Reality (ISMAR)}.\hskip 1em plus 0.5em minus 0.4em\relax IEEE, 2022, pp. 499--507.

\bibitem{yu2021pixelnerf}
A.~Yu, V.~Ye, M.~Tancik, and A.~Kanazawa, ``pixelnerf: Neural radiance fields from one or few images,'' in \emph{Proceedings of the IEEE/CVF Conference on Computer Vision and Pattern Recognition}, 2021, pp. 4578--4587.

\bibitem{yuan2022nerf}
Y.-J. Yuan, Y.-T. Sun, Y.-K. Lai, Y.~Ma, R.~Jia, and L.~Gao, ``Nerf-editing: geometry editing of neural radiance fields,'' in \emph{Proceedings of the IEEE/CVF Conference on Computer Vision and Pattern Recognition}, 2022, pp. 18\,353--18\,364.

\bibitem{yue2005similarity}
J.~C. Yue and M.~K. Clayton, ``A similarity measure based on species proportions,'' \emph{Communications in Statistics-theory and Methods}, vol.~34, no.~11, pp. 2123--2131, 2005.

\bibitem{zhang2023nerf}
J.~Zhang, F.~Zhang, S.~Kuang, and L.~Zhang, ``Nerf-lidar: Generating realistic lidar point clouds with neural radiance fields,'' \emph{arXiv preprint arXiv:2304.14811}, 2023.

\bibitem{zhang2018unreasonable}
R.~Zhang, P.~Isola, A.~A. Efros, E.~Shechtman, and O.~Wang, ``The unreasonable effectiveness of deep features as a perceptual metric,'' in \emph{Proceedings of the IEEE conference on computer vision and pattern recognition}, 2018, pp. 586--595.

\bibitem{zhu2022nice}
Z.~Zhu, S.~Peng, V.~Larsson, W.~Xu, H.~Bao, Z.~Cui, M.~R. Oswald, and M.~Pollefeys, ``Nice-slam: Neural implicit scalable encoding for slam,'' in \emph{Proceedings of the IEEE/CVF Conference on Computer Vision and Pattern Recognition}, 2022, pp. 12\,786--12\,796.

\end{thebibliography}
}
\end{document}